\documentclass[12pt]{iopart}
\usepackage{iopams}
\usepackage{multirow}
\usepackage{afterpage} 
\usepackage{arydshln} 
\usepackage{url}
\usepackage{hyperref}

\usepackage{graphicx}
\usepackage[compress,sort,numbers]{natbib}

\newcommand{\newblock}{\hskip .11em\@plus.33em\@minus.07em}

\begin{document}
\begin{flushright}
FERMILAB-PUB-24-0945-CSAID
\end{flushright}
\title[Neural Architecture Codesign for Fast Physics Applications]{Neural Architecture Codesign for Fast Physics Applications}


\author{Jason Weitz$^{1}$\footnote{These authors contributed equally to this work.}, Dmitri Demler$^{1}$\footnotemark[1], 
Luke McDermott$^1$, Nhan Tran$^{2,3}$, and Javier Duarte$^1$}

\address{$^1$ Department of Physics, University of California San Diego, La Jolla, CA 92093, USA}
\address{$^2$ Fermi National Accelerator Laboratory, Batavia, IL 60510, USA}
\address{$^3$ McCormick School of Engineering, Northwestern University, Evanston, IL 60208, USA}

\ead{jdweitz@ucsd.edu}

\begin{abstract}
We develop a pipeline to streamline neural architecture codesign for physics applications to reduce the need for ML expertise when designing models for novel tasks.
Our method employs neural architecture search and network compression in a two-stage approach to discover hardware efficient models.
This approach consists of a global search stage that explores a wide range of architectures while considering hardware constraints, followed by a local search stage that fine-tunes and compresses the most promising candidates.
We exceed performance on various tasks and show further speedup through model compression techniques such as quantization-aware-training and neural network pruning.
We synthesize the optimal models to high level synthesis code for FPGA deployment with the \texttt{hls4ml} library.
Additionally, our hierarchical search space provides greater flexibility in optimization, which can easily extend to other tasks and domains.
We demonstrate this with two case studies: Bragg peak finding in materials science and jet classification in high energy physics, achieving models with improved accuracy, smaller latencies, or reduced resource utilization relative to the baseline models.

\end{abstract}

\submitto{\MLST}
\maketitle

\section{Introduction}

Deep learning has proven to be a powerful tool for tackling complex problems across many scientific domains, from materials science to particle physics. 
However, designing an optimal neural network architecture given a given task, computational platform, and latency or resource constraints remains a significant challenge, often requiring extensive trial and error which is time-consuming, computationally expensive, and may not result in the best possible performance.
Moreover, keeping pace with the rapid advancements in deep learning techniques can be daunting for researchers whose primary expertise lies outside the field of machine learning.
To alleviate this problem, neural architecture search (NAS)~\cite{NAS} can greatly accelerate the design and deployment of deep learning models while also reducing the burden on domain experts to stay abreast of the latest machine learning developments.

NAS aims to automate the design of neural networks, enabling the discovery of high-performing architectures tailored to specific tasks and constraints.
By systematically exploring a predefined search space of potential architectures, NAS algorithms can identify novel network designs that match or exceed the performance of hand-crafted models.
This is achieved through a combination of search strategies, such as evolutionary algorithms~\cite{evolutionary} and efficient architecture evaluation methods~\cite{efficient}, like weight sharing or surrogate models.
The key advantage of NAS is its ability to find optimal architectures for a given problem without relying on human expertise or trial-and-error, thus saving time and resources while potentially yielding better results.
However, optimizing for multiple objectives, such as accuracy and computational complexity, to find the best models along a Pareto-optimal front is still an area of active research~\cite{shariatzadeh2023surveymultiobjectiveneuralarchitecture}.


\begin{figure}[p]
    \centering
    \includegraphics[width=0.75\linewidth]{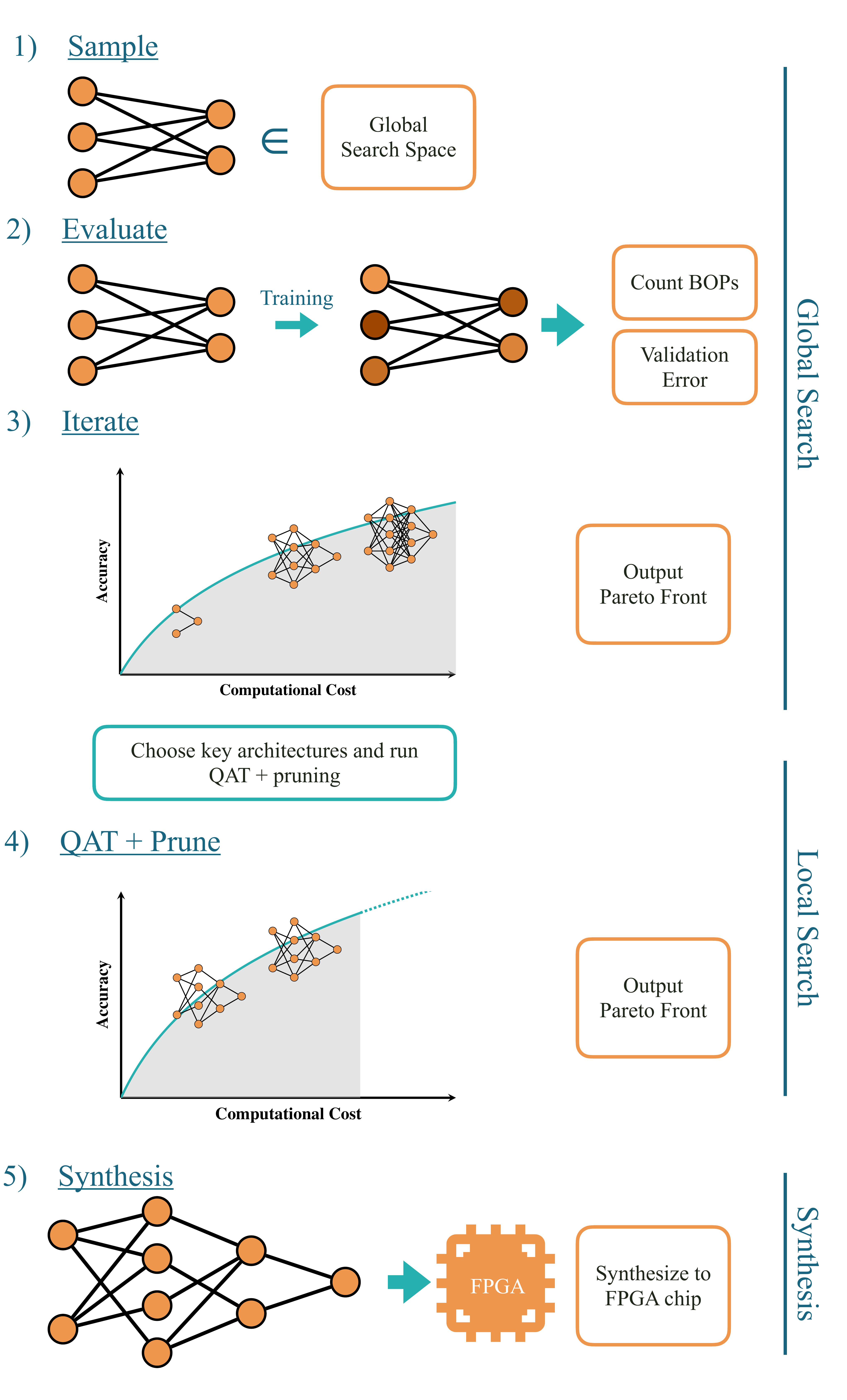}
    \caption{\label{fig:FullPipeline}Full pipeline methodology of neural architecture codesign containing the two stages: global and local search.
    Global search stage explores a wide range of architectures and local search further fine-tunes hyperparameters and applies compression techniques.
    We then synthesize the optimal models to high level synthesis code for FPGA deployment.}
\end{figure}

In this paper, we introduce \emph{neural architecture codesign (NAC)}, an extension of NAS that optimizes neural architectures for both task performance and hardware efficiency.
We present a general-purpose NAC framework that can be adapted to a wide range of tasks and domains.
Our approach, illustrated in Fig.~\ref{fig:FullPipeline}, consists of a two-stage optimization process based on the once-for-all methodology~\cite{cai2020once}---a global search stage that explores a diverse set of architectures while considering hardware constraints, followed by a local search stage that fine-tunes and compresses the most promising candidates.
We employ a flexible and modular search space that allows for the incorporation of domain-specific knowledge and constraints.
Additionally, we introduce techniques for efficient evaluation of candidate architectures during the search process, balancing the accuracy of performance estimates with computational cost.
To make our NAC pipeline accessible to researchers across scientific fields, we encapsulate state-of-the-art techniques within a user-friendly framework that integrates popular open-source packages.
These include Brevitas~\cite{brevitas} and QKeras~\cite{QAT} for quantization-aware training, Optuna~\cite{Optuna} for hyperparameter optimization, and \texttt{hls4ml}~\cite{hls4ml} for automated deployment of optimized models on FPGAs.
We present the effectiveness of our NAC methodology in two distinct scientific tasks, showing its ability to discover models that outperform hand-crafted architectures and being more efficient when deployed on hardware.

To demonstrate the versatility and effectiveness of our NAC framework, we present two case studies from distinct physics domains.
The first study focuses on BraggNN~\cite{BraggNN}, a deep learning model for fast X-ray Bragg peak analysis in high-energy diffraction microscopy experiments in materials physics.
The second case study explores the use of the deep sets~\cite{zaheer2017deep,Komiske:2018cqr,odagiu2024sets} neural architecture for the classification of high-energy particle jets in particle physics, known as \emph{jet tagging}~\cite{Mondal:2024nsa}.
Our NAC framework software is publicly available~\cite{dmitri_demler_2025_14618350}.

\section{Related Work}

\subsection{Neural Architecture Search}

NAS aims to optimize the structure of neural networks for specific tasks and objectives.
This includes searching over network sizes or even constructing completely different model classes.
There are three critical components: \textit{search space}, \textit{search strategy}, and \textit{architecture evaluation}.
The search space determines the potential architectures that can be sampled~\cite{search_space1}.
While a narrow search space can be heavily biased, a large one is extremely difficult to properly explore, necessitating a delicate balance.
This space is explored by sampling architectures and evaluating them across our metrics.
Instead of evaluating high-cost objectives, such as network performance, researchers often use a proxy for such objectives like partial training or even zero-cost methods~\cite{zero_cost_nas}.
After validating the candidate architecture, the search strategy will update its beliefs and sample again.
Various strategies for this exist, such as Bayesian optimization (BO), evolutionary algorithms, or reinforcement learning, with differing strengths and weaknesses.
For example, BO methods struggle with many categorical hyperparameters due to the large number of combinatorial possibilities, prompting the use of genetic algorithms instead~\cite{cai2020once}.
However, BO, specifically tree-structured Parzen estimators (TPEs)~\cite{TPE}, performs exceptionally well on continuous hyperparameter optimization tasks where sample efficiency is important~\cite{Optuna}.
In this paper, we use the non-dominated sorting genetic algorithm (NSGA-II)~\cite{NSGA}.
NSGA-II maintains a population of candidate architectures and evolves them over multiple generations using genetic operations like mutation and crossover.
The algorithm ranks the candidates based on their performance and diversity, promoting solutions that are both high-performing and diverse in their architectural choices.
TPE is used for the continuous hyperparameters in training optimization.

\subsection{Model Compression}

In addition to finding an efficient model configuration, models can be further optimized through neural network pruning and quantization~\cite{liang2021pruning}. 
Pruning aims to remove a model's superfluous parameters~\cite{pruning_survey}, while quantization reduces the number bits needed to represent them.

Structured pruning removes weights associated to larger structures in the network, such as neurons, channels, or attention heads.
Removing large structures can be seen as reducing the dimensionality of the weight tensor, by decreasing the amount of rows or columns in the matrix. 
On the other hand, unstructured pruning removes individual weights with no specific structure requirements.
This leads to sparse matrices of the original dimension, which can often limit gains in actual inference speed on GPUs despite the reduced number of parameters.
However, on more versatile or flexible hardware like FPGAs and CPUs, unstructured pruning can provide significant speed up with a negligible drop in performance~\cite{sui2023hardware}.
While structured pruning provides definite inference time improvements on general hardware, it can lead to a larger decline in performance.
To get around this, newer hardware supports $N{:}M$ or mixed sparsity~\cite{nvidiaNM}, such as Nvidia's A100 that supports 2:4 sparsity, which alleviates the need to prune entire rows or columns as done with dropping neurons or filters.
Therefore, the choice of pruning algorithm is closely tied to the target hardware for deployment.

Quantization reduces the number of bits needed to represent weights or activations.
Like pruning, quantization can be done post-training (PTQ) or during, with quantization-aware training (QAT)~\cite{QAT}.
With QAT, the weights are quantized on the forward pass, but use full-precision gradients on the backward pass, allowing for further fine-tuning of the low-bit representations.
The effectiveness of quantization is heavily dependent on hardware support of low-bit data types.
For the current study, we focus on deployment on FPGAs, which support sparse operations and a wide range of reduced precision data types; thus we use unstructured pruning with QAT.

\subsection{Hardware Implementation and Synthesis}

Many experiments require fast inference times, and FPGAs are often used to achieve this goal.
FPGAs offer several advantages over traditional computing architectures.
They provide faster and more efficient processing compared to CPUs, while also allowing for more flexibility and customization than ASICs. 
This makes FPGAs well-suited for implementing machine learning models in scientific applications that demand high throughput and low latency.

However, deploying machine learning models on FPGAs can be challenging due to their fixed architecture and limited resources compared to GPUs.
To address this, tools like \texttt{hls4ml}~\cite{hls4ml} have been developed to streamline the process of synthesizing ML models into FPGA firmware.
\texttt{hls4ml} is an open-source library that translates models from common open-source machine learning frameworks, like TensorFlow and PyTorch, into high-level synthesis (HLS) code in C++.
This HLS code can then be synthesized into FPGA firmware using commercial tools like Xilinx Vivado.
By automating much of the process, \texttt{hls4ml} significantly lowers the barrier to entry for deploying ML on FPGAs.

\texttt{hls4ml} supports a variety of layer types and network architectures, and provides configuration options to tune the implementation for the specific use case and target FPGA.
It also supports techniques like quantization and pruning to reduce the resource usage and latency of the synthesized model.
While \texttt{hls4ml} was initially developed for particle physics applications, it is broadly applicable to other scientific domains that can benefit from fast ML inference on FPGAs~\cite{Deiana:2021niw}.

\section{Method}

Our proposed NAC framework consists of a two-stage optimization process: a \textit{global search} stage and a \textit{local search} stage.
The global search stage explores a wide range of architectures within a predefined search space to identify promising candidate models.
The local search stage fine-tunes the hyperparameters and compresses these candidate models to further improve their performance and optimize them for the specific task at hand.
These stages are followed by a hardware implementation step for FPGA use, in which the resulting model is synthesized for hardware estimation and deployment.

\subsection{Global Search}

\subsubsection{Search space}

The search space defines the set of possible architectures that can be explored during the optimization process.
We design a flexible and modular search space that can be easily adapted to various tasks and domains.


The search space consists of a series of blocks, each representing a specific type of neural network layer or operation.
These blocks can include convolutional layers, fully-connected layers, attention mechanisms, pooling operations, normalization, and activation functions.
Each block has associated hyperparameters that control its behavior, such as the number of filters in a convolutional layer, the number of neurons in a fully-connected layer, or the type of activation function.

\subsubsection{Search and Evaluation}

The search strategy is responsible for efficiently exploring the search space and identifying promising candidate architectures.
In our framework, we employ a multi-objective optimization algorithm that considers both the performance of the models on the target task and their computational efficiency.
This approach ensures that the resulting architectures not only achieve high accuracy but also meet the resource constraints of the target platform, such as inference time or memory usage.
We implement NSGA-II for this multi-objective optimization in the global search.



In addition to the performance metrics, we also evaluate the computational efficiency of the candidate architectures using bit operations (BOPs) or the actual inference time on the target hardware platform\footnote{More information on the BOPs calculation can be found in \ref{sec: Bit Operations Calculation}.}. 
By incorporating these metrics into the optimization process, we can guide the search towards architectures that strike a balance between performance and efficiency.

Throughout the search process, we maintain an archive of the most promising architectures discovered so far.
This serves as a valuable resource for the subsequent local search stage, providing a diverse set of high-quality candidate solutions that can be further refined and optimized.
Models are selected based on their respective performance and categorized according to their BOPs.

\subsection{Local Search}

\subsubsection{Training Optimization}
After finding multiple high performing models with global search, we now perform a hyperparameter optimization of the training procedure via tree-structured Parzen estimation.
These hyperparameters consist of learning rate, learning rate schedule, weight decay, etc.


\subsubsection{Model compression}
To improve the efficiency even further, QAT can be paired with neural network pruning.
The bit precisions used are 4, 8, 16, and 32.
For each bit precision, iterative magnitude-based unstructured pruning is performed with QAT using Brevitas~\cite{brevitas} inside the inner loop, removing 20\% of the parameters each iteration, for a total of 20 iterations.
This ultimately produces models ranging from 0 to 99\% sparsity.
The models are chosen with the goal of lowest bit precision, highest sparsity, and best performance.

\subsection{Model FPGA Synthesis}
With the optimal models found after the two-stage optimization process, the architectures are now synthesized for hardware deployment.
In this process, key configurability options including precision, strategy, and reuse factor are chosen for optimization.
The reuse factor indicates the number of multipliers used for multiplication operations within the network's computation of values.
Increasing the reuse factor reduces resource utilization while increasing latency and decreasing throughput.
The precision option configures the bit precision of the weights in the model, which can be altered at the layer level.
One of two strategies, ``latency'' or ``resource,'' can be selected, which optimize for lower latency or lower resource utilization, respectively.
Using \texttt{hls4ml} version 0.8.0~\cite{fastml_hls4ml} and AMD Vivado 2020.1, the respective model is translated from a machine learning framework to HLS code for deployment.


\section{Bragg Peak Case Study}


\begin{figure}
    \centering
    \caption{\label{fig:Bragg Peak Visual}Bragg peak NAC visualization.
    The input consists of 11$\times$11 pixel patches centered on Bragg peaks from X-ray diffraction patterns.
    The neural network predicts the $x$ and $y$ coordinates of the peak center within each patch.}
    \includegraphics[width=0.9\textwidth]{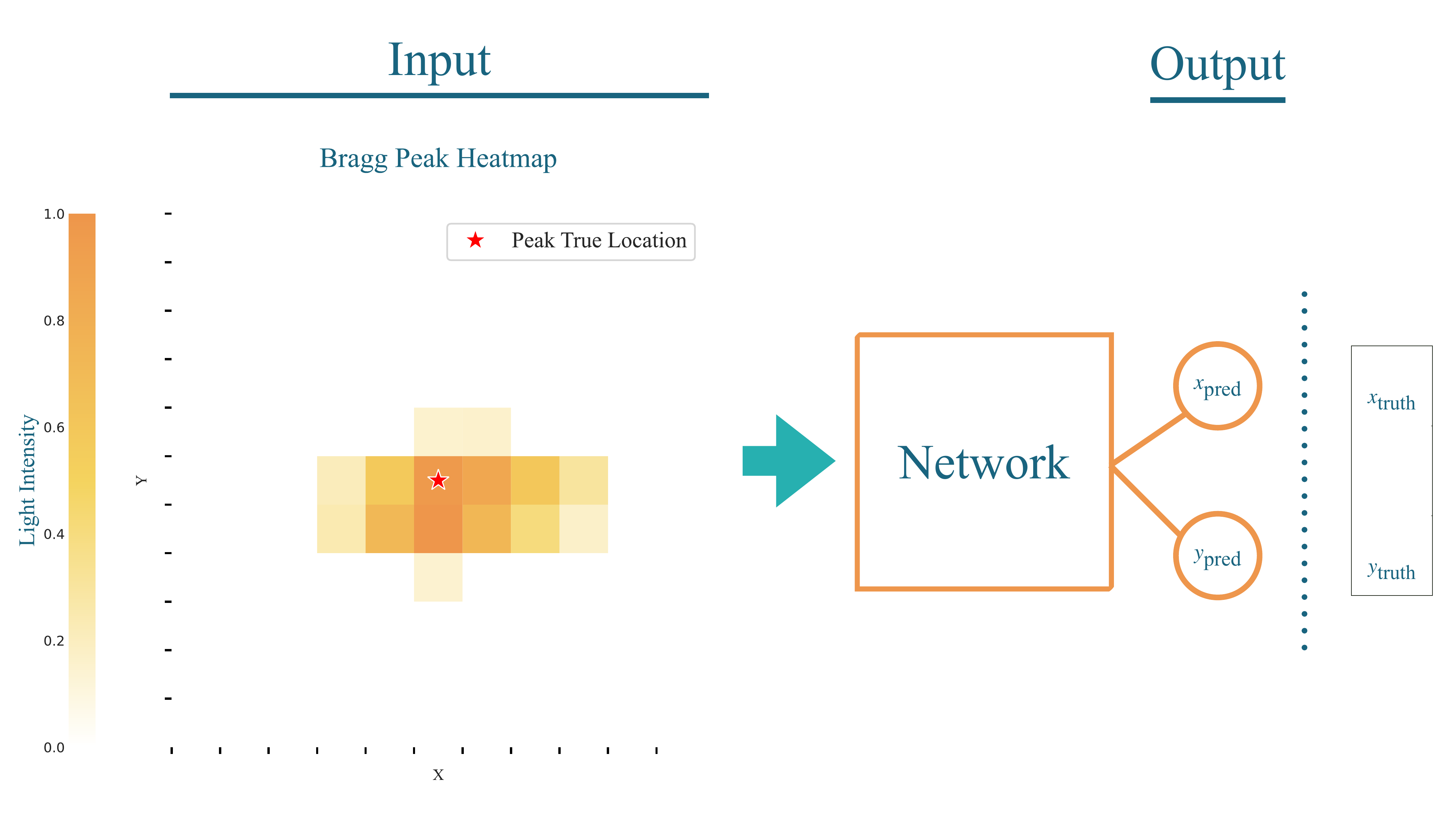}
\end{figure}

BraggNN is a materials science deep learning model developed for fast X-ray Bragg peak analysis in high-energy diffraction microscopy (HEDM) experiments~\cite{BraggNN}.
HEDM is a technique used to characterize the microstructure and micromechanical state of polycrystalline materials~\cite{park2017far}.
A key step in the data analysis pipeline is fitting Bragg peaks to a pseudo-Voigt profile, which is computationally expensive and can be a bottleneck for real-time analysis.
BraggNN aims to accelerate this process by using a convolutional neural network (CNN) instead to directly predict the peak center positions.

The input data to BraggNN are small 11$\times$11 pixel patches cropped around each Bragg peak, depicted in Fig.~\ref{fig:Bragg Peak Visual}.
The dataset contains approximately 70,000 such patches from a scan of a gold sample.
The small size and sparsity of the input patches make the dataset well-suited for deployment on edge devices like FPGAs that have limited on-chip memory and benefit from low-precision computation.
OpenHLS~\cite{OpenHLS} further optimized BraggNN for deployment on FPGAs demonstrating significant speed-ups for real-time analysis.

\subsection{Method Adaptations}

\begin{figure}
    \caption{\label{BragNN Pipeline}Our automated pipeline for neural architecture search for the Bragg peak dataset.
    Yellow components are human inputs, white are outputs, and orange are search processes. 
    The right side demonstrates the template of each candidate architecture in our search space.
    Each subcomponent of the blocks also contains the hyperparameters to optimize.}
    \includegraphics[width=\textwidth]{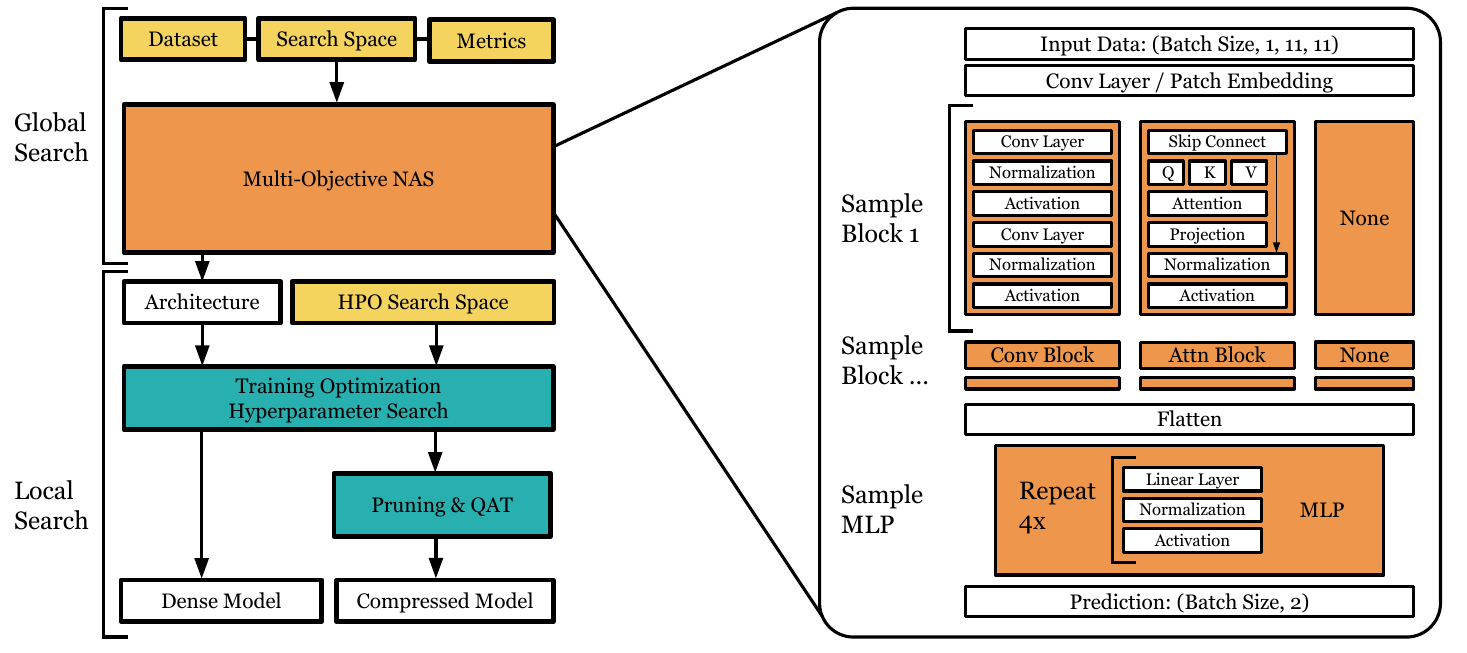}
\end{figure}

We used the same general two-stage neural architecture search strategy outlined in Section 3.
In the global search stage, we utilized a similar search space composed of convolutional, attention, and fully-connected blocks as seen in Fig.~\ref{BragNN Pipeline}.
However, we restricted the space to small architectures to ensure compatibility with the Xilinx Virtex UltraScale+ U250 (\texttt{xcu250-figd2104-2L-e}) target FPGA.
For the local search stage, we again used HPO to fine-tune the training hyperparameters. 

\subsection{Bragg Peak Results}
We evaluated our optimized models on the gold diffraction dataset.
Tables~\ref{tab:Bragg model_comparison1} and~\ref{tab:Bragg model_comparison2} provide a comprehensive comparison of our optimized models to the original BraggNN implementation in terms of performance, BOPs, parameters, FPGA resources, and latency, respectively. The resuse factor is a hyperparameter that we fix but can be varied in future studies.

\begin{table}[ht]
\caption{\label{tab:Bragg model_comparison1}Bragg Peak Model comparison with respect to distance, MegaBOPs, and parameters.}
\centering
\footnotesize
\setlength{\tabcolsep}{6pt}
\begin{tabular}{@{}c*{3}{c}@{}}
\br
Model & Mean Distance [pixels] & MegaBOPs & Parameters \\
\mr
BraggNN & 0.202 & 34,540 & 45,274 \\
Large   & \textbf{0.201} & 5,861  & 92,438 \\
Medium  & 0.208 & 1,447  & 26,402 \\
Small   & 0.211 & 881    & 23,788 \\
Tiny    & 0.227 & \textbf{800}    & \textbf{23,094} \\
\br
\end{tabular}
\end{table}


\begin{table}[ht]
\caption{\label{tab:Bragg model_comparison2}Comparison of different Bragg peak models' latency, initiation interval (II) based on clock cycles (cc), and hardware resource utilization.
DSP and BRAM are not listed as they are not utilized (0\%).
All models are quantized to 8 bits.
A reuse factor of 4 is used.
The baseline BraggNN model is not listed, as its attention block cannot be currently synthesized.}
\centering
\scriptsize
\setlength{\tabcolsep}{4pt}
\begin{tabular}{@{}p{1.5cm}*{5}{c}@{}}
\br
Model & Latency [$\mu$s] (cc) & II [$\mu$s] (cc) & DSP & LUT [\%] & FF \\
\mr
Large   & 8.56 (1711) & 1.82--8.48 (365--1696) & 6433 (52.35\%) & 181,298 (10.49\%) & 124,887 (3.61\%) \\
Medium  & 4.94 (987)   & \textbf{1.82--4.85 (365--970)}  & 2697 (21.95\%) & 47,903 (2.77\%)   & 54,937 (1.59\%) \\
Small   & 4.92 (984)   & \textbf{1.82--4.85 (365--970)}  & \textbf{290 (2.36\%)}   & 121,941 (7.06\%)  &\textbf{ 34,765 (0.70\%)} \\
Tiny    & \textbf{4.88 (975)}   & \textbf{1.82--4.85 (365--970)}  & 3148 (25.62\%) & \textbf{40,801 (2.36\%)}   & 36,360 (1.05\%) \\
\br
\end{tabular}
\end{table}

\begin{figure}[htpb]
\centering
\includegraphics[width=0.48\textwidth]{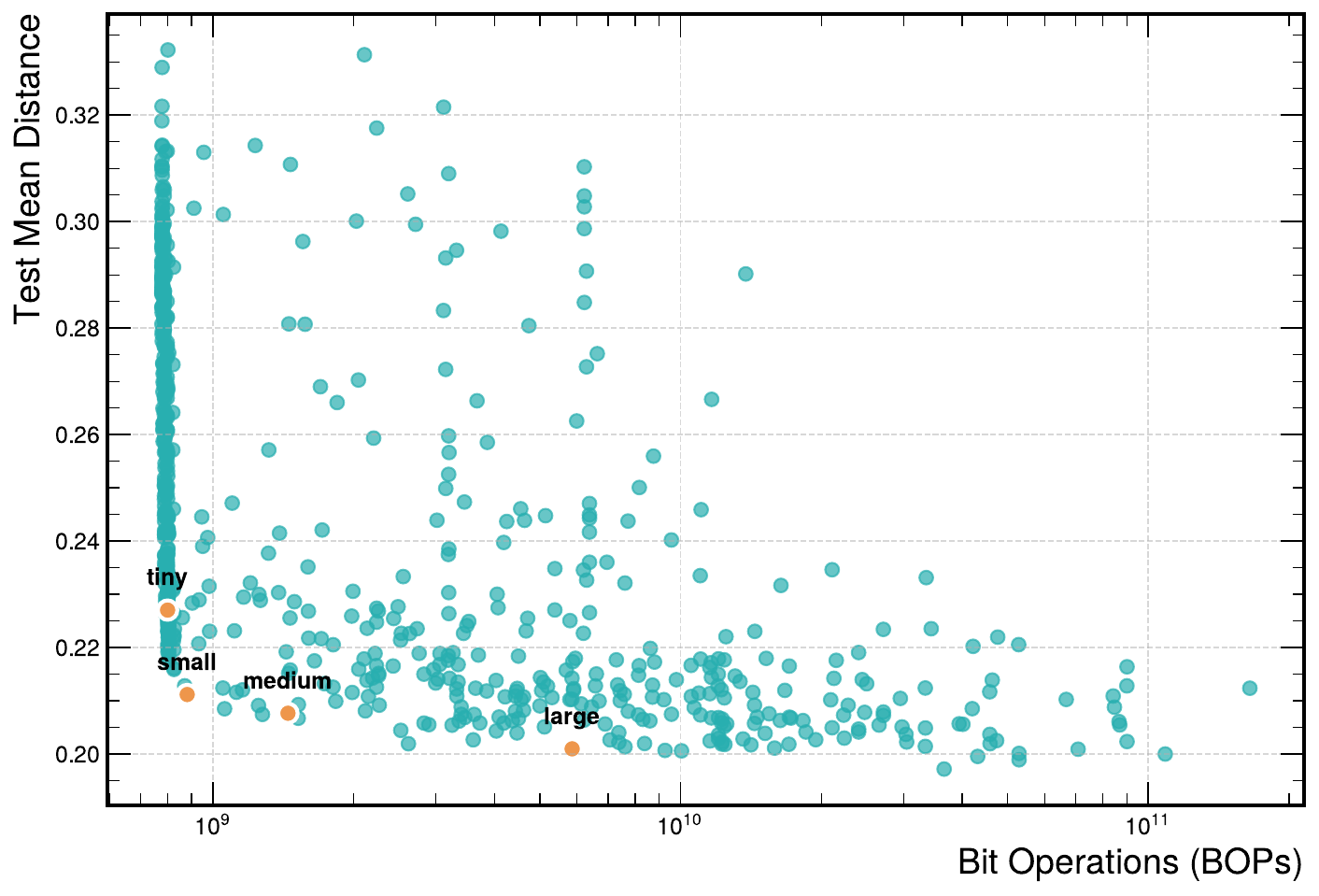}
\includegraphics[width=0.48\textwidth]{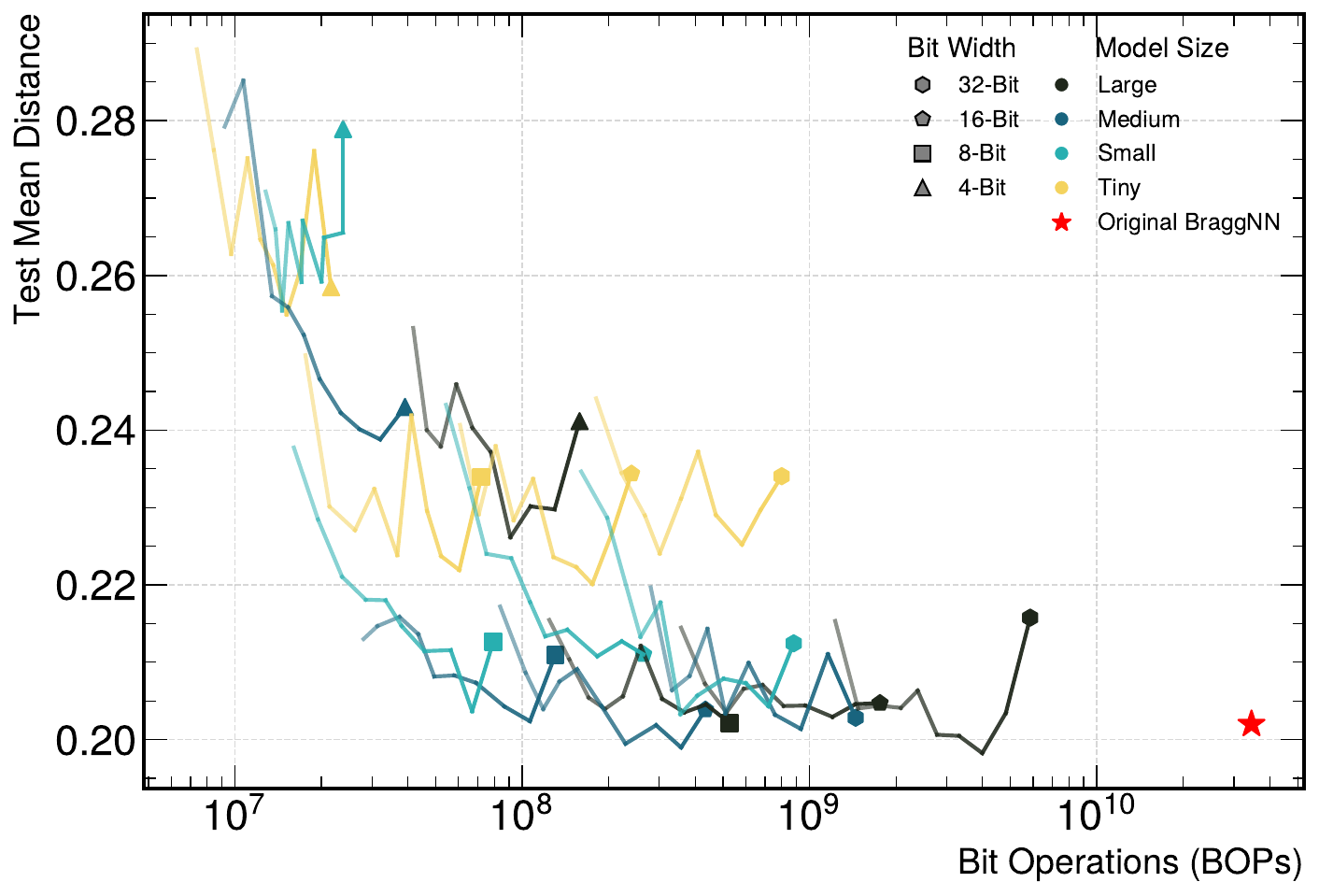}
\caption{\label{fig:bragg-pareto-front}Bragg peak dataset Pareto-optimal front for global (left) and local search (right).
The global search contains 1000 trials with the selected models in orange.
For the local search, each chosen model is quantized to various bit precisions, and then pruned by 20\% for 20 iterations indicated by increasing opacity.
The first 10 iterations are displayed for visualization purposes.}
\end{figure}

In the BraggNN case study, our large model slightly improved the mean distance while achieving a 5.9$\times$ reduction in BOPs, albeit with a 2$\times$ increase in parameters.
This showcases the imperfection of efficiency metrics that aim to approximate the actual latency and resource utilization when synthesized to the FPGA.
The small model has a better balance: with only a 3\% larger mean distance but with a 39.2$\times$ decrease in BOPs and a 2.90$\times$ decrease in parameters. 
Full model architectures can be accessed in Appendix Table~\ref{tab:braggnn_models}.
All models can be pruned to 80\% sparsity (8 iterations) with less than a 10\% drop in performance, depicted in Fig.~\ref{fig:bragg-pareto-front}.
BraggNN uses a non-local attention block which we find is unnecessary for performance.

When synthesized for FPGA deployment, our optimized models demonstrate significant improvements in latency and resource utilization.
Our small model achieves a latency of 4.920 $\mu$s with a 1.825 $\mu$s initialization interval, using 2.36\% of the available DSPs, 7.06\% of the LUTs, and 1.05\% of the flip-flops on the target FPGA (Table~\ref{tab:Bragg model_comparison2}).

The performance of the BraggNN models during the global and local searches can be seen in Fig.~\ref{fig:bragg-pareto-front}. Each point indicates a unique trained model architecture, demonstrating the exploration and refinement of the search process.
The global search stage evaluated 1000 model architectures, utilizing 1 NVIDIA 4090 GPU for 96 hours.
The local search stage trained 4 models at 4 precisions, with iterative magnitude pruning for 20 iterations, with 100 epochs per iteration. 4 NVIDIA 3090 GPUs were used in parallel for 29 hours for this stage.

\section{Jet Classification Case Study}

\begin{figure}[htpb]
    \centering
    \includegraphics[width=0.9\linewidth]{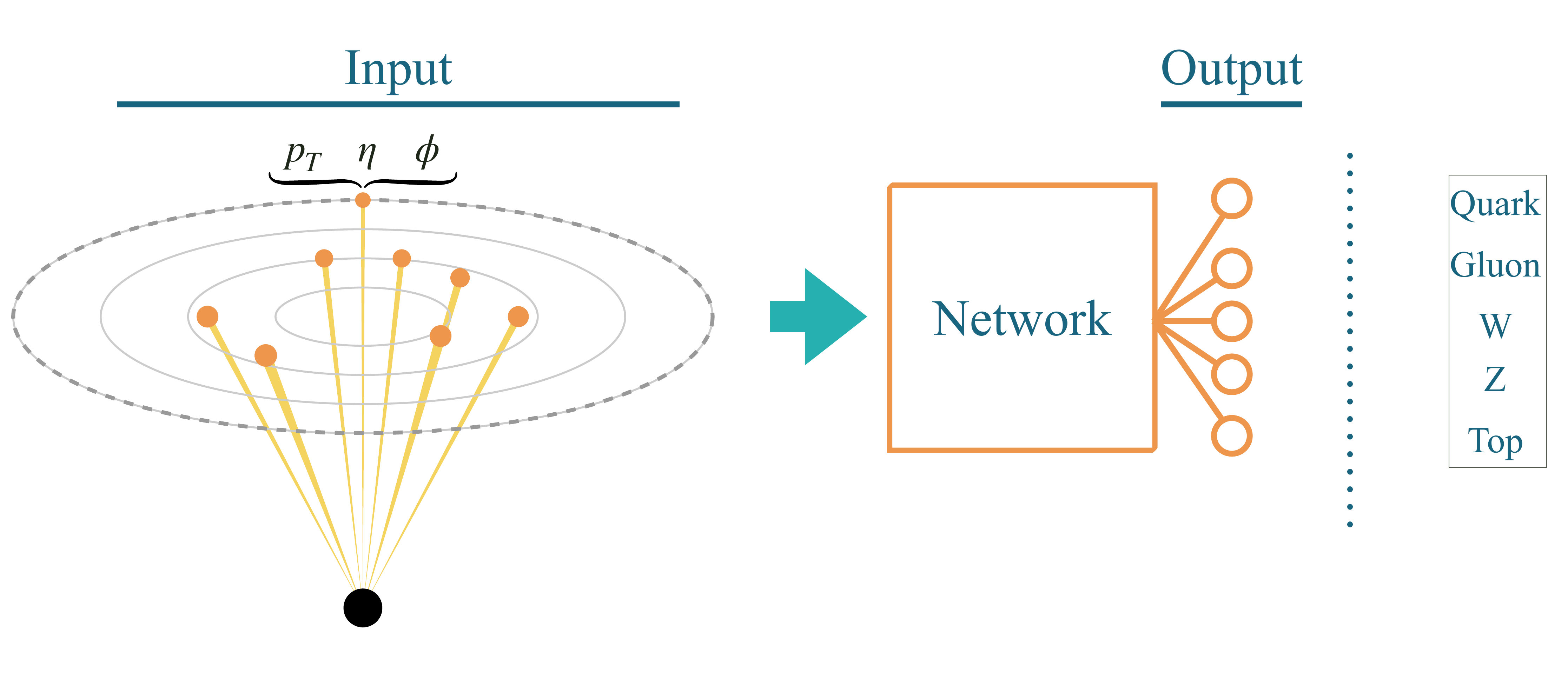}
    \caption{\label{fig: Deepsets Flow Chart}Particle jet NAC visualization.
    The input consists of sets of up to 8 constituent particles from a jet, with the transverse momentum ($p_\mathrm{T}$), pseudorapidity ($\eta$), and azimuthal angle ($\phi$) known for each particle.
    The neural network predicts the origin of the jet (light quark, gluon, W boson, Z boson, or top quark).}
\end{figure}

In particle physics, jets are collimated sprays of particles that originate from the decay of heavy particles like top quarks, W bosons, or Z bosons.
Identifying the type of particle that initiated a jet, a task known as jet tagging, is crucial for many particle physics analyses as it provides valuable information about the underlying physics processes.
Jet tagging can be formulated as a set-based problem, where each jet is represented as a set of its constituent particles, and the goal is to classify the jet based on the properties of these particles, as depicted in Fig.~\ref{fig: Deepsets Flow Chart}.

Deep sets is a neural network architecture designed to operate on sets of objects, where the order of the elements in the set does not matter~\cite{zaheer2017deep}.
This permutation-invariant property makes deep sets particularly suitable for tasks involving unordered sets of variable size.
The key idea behind deep sets is to apply a shared neural network, $\phi$, to each element of the input set independently.
$\phi$ maps each input element to a high-dimensional representation.
These individual representations are then pooled using a permutation-invariant function such as sum, mean, or max operations, resulting in a fixed-length embedding that captures the relevant information from the entire set.
Next, another neural network, $\rho$, is applied to the pooled embedding to produce the desired classification output.

Deep sets are well-suited for this task because they can handle variable-length sets of particles and are invariant to the order in which the particles are presented, which is a desirable property since the order of particles within a jet is not physically meaningful.

\subsection{Method Adaptation} 

In the global search stage, we explored a wide range of hyperparameters to find the optimal configuration for the deep sets architecture.
These hyperparameters included the dimensionality of the latent space after applying the permutation-invariant pooling operation, the choice of pooling operation (sum, mean, or max), and the architectures of the $\phi$ and $\rho$ MLPs.
The $\phi$ MLP is applied independently to each element of the input set, while the $\rho$ MLP operates on the pooled embeddings.

In the local search stage, we used HPO to fine-tune the training hyperparameters, such as learning rate, batch size, and regularization strength.
Since inference speed is critical for jet tagging in real-time particle physics experiments, we also used model compression techniques like quantization and pruning during this stage to reduce the model size and computational cost while maintaining high classification accuracy. 

For the FPGA implementation, we used a custom branch of \texttt{hls4ml}~\cite{customhls4ml_branch} that allows for more flexible reuse and parallelization factors for the deep sets architecture.
This enabled us to better optimize the model for the specific target FPGA, balancing resource utilization and inference speed.
The target FPGA is the Xilinx Virtex UltraScale+ VU13P (\texttt{xcvu13p-flga2577-2-e}).

\subsection{Particle Jet Results}

Tables~\ref{tab:Deepsets model_comparison1} and~\ref{tab:Deepsets model_comparison2} provide a comprehensive comparison of our optimized models with the original deep sets implementation for the particle jet classification task.

\begin{table}[ht]
\caption{\label{tab:Deepsets model_comparison1}Comparison of the particle jet models' accuracy, MegaBOPs, and parameters.}
\centering
\footnotesize
\setlength{\tabcolsep}{6pt}
\begin{tabular}{@{}c*{3}{c}@{}}
\br
Model & Accuracy [\%] & MegaBOPs & Parameters \\
\mr
Baseline & 64.0 & 12.10 & 3,461 \\
Large     & \textbf{66.55} & 5.23  & 7,535 \\
Medium    & 65.06 & 1.68  & 2,813 \\
Small     & 63.06 & 0.75  & 815 \\
Tiny      & 61.16 & \textbf{0.40}  & \textbf{573} \\
\br
\end{tabular}
\end{table}

\begin{table}[ht]
\caption{\label{tab:Deepsets model_comparison2}Comparison of the particle jet models'  latency, initiation interval (II), and hardware utilization, including arithmetic DSPs, flip flops (FF)s, and block RAM (BRAM).
All models are quantized to 8 bits.
A reuse factor of 2 is used.}
\centering
\scriptsize
\setlength{\tabcolsep}{4pt}
\begin{tabular}{@{}p{1.5cm}*{6}{c}@{}}
\br
Model & Lat. [ns] (cc) & II [ns] (cc) & DSP & LUT & FF & BRAM \\
\mr
Baseline & 95 (19) & 15 (3) & 626 (5.1\%) & 386,294 (22.3\%) & 121,424 (3.5\%) & 4 (0.1\%) \\
Large     & 135 (27) & 15 (3) & 2,458 (20.0\%) & 337,172 (19.51\%) & 139,905 (4.05\%) & 4 (0.1\%) \\
Medium    & 110 (22) & 15 (3) & 548 (4.46\%) & 130,426 (7.55\%) & 49,326 (1.43\%) & 4 (0.1\%) \\
Small     & 105 (21) &\textbf{ 10 (2)} &\textbf{ 302 (2.46\%)} & 53,398 (3.09\%) & 28,057 (0.81\%) & 4 (0.1\%) \\
Tiny      & \textbf{70 (14)}  & \textbf{10 (2)} & 321 (2.61\%) & \textbf{32,884 (1.90\%)} &\textbf{ 15,918 (0.46\%)} & 4 (0.1\%) \\
\br
\end{tabular}
\end{table}

\begin{figure}[htpb]
\includegraphics[width=0.48\textwidth]{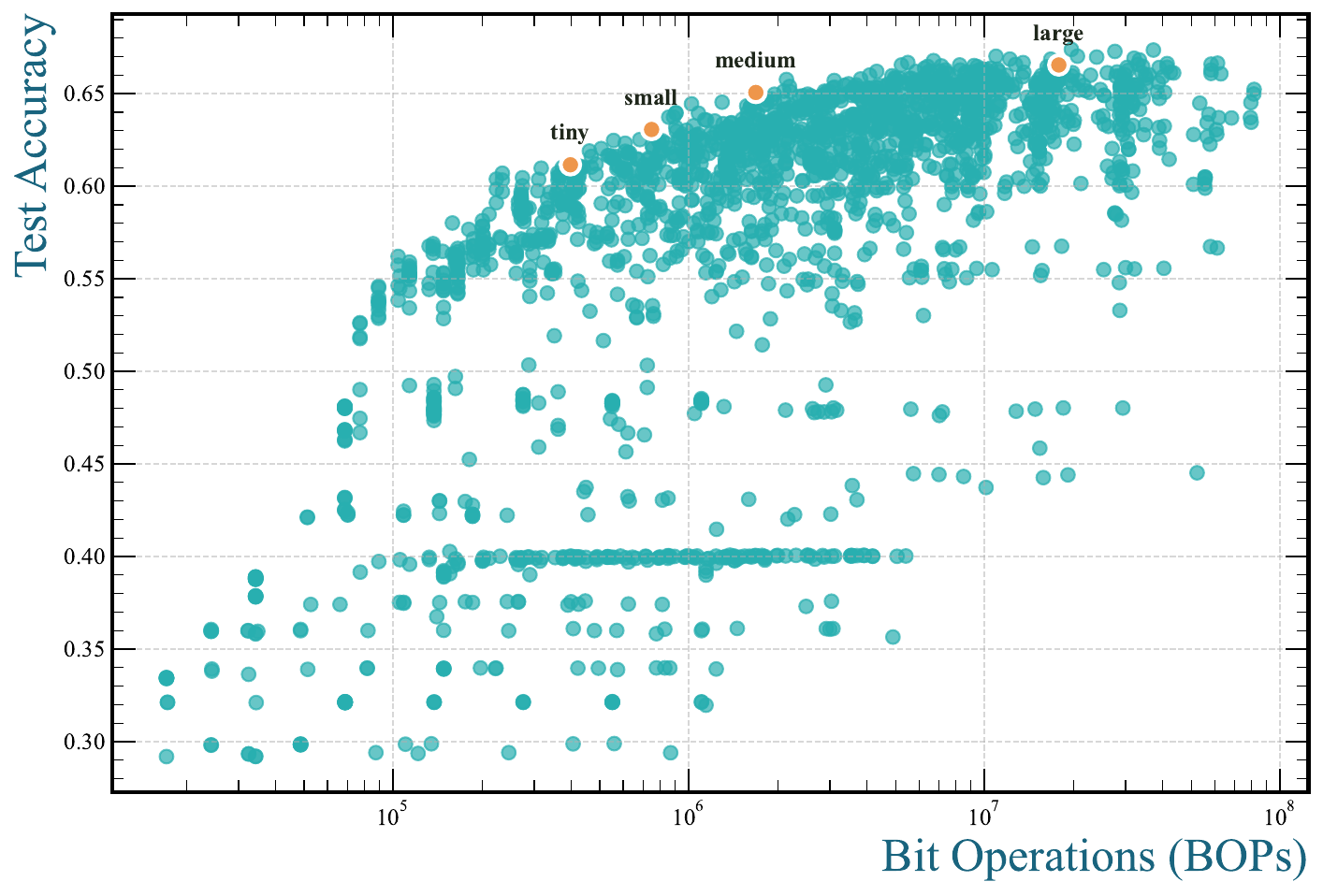}
\includegraphics[width=0.48\textwidth]{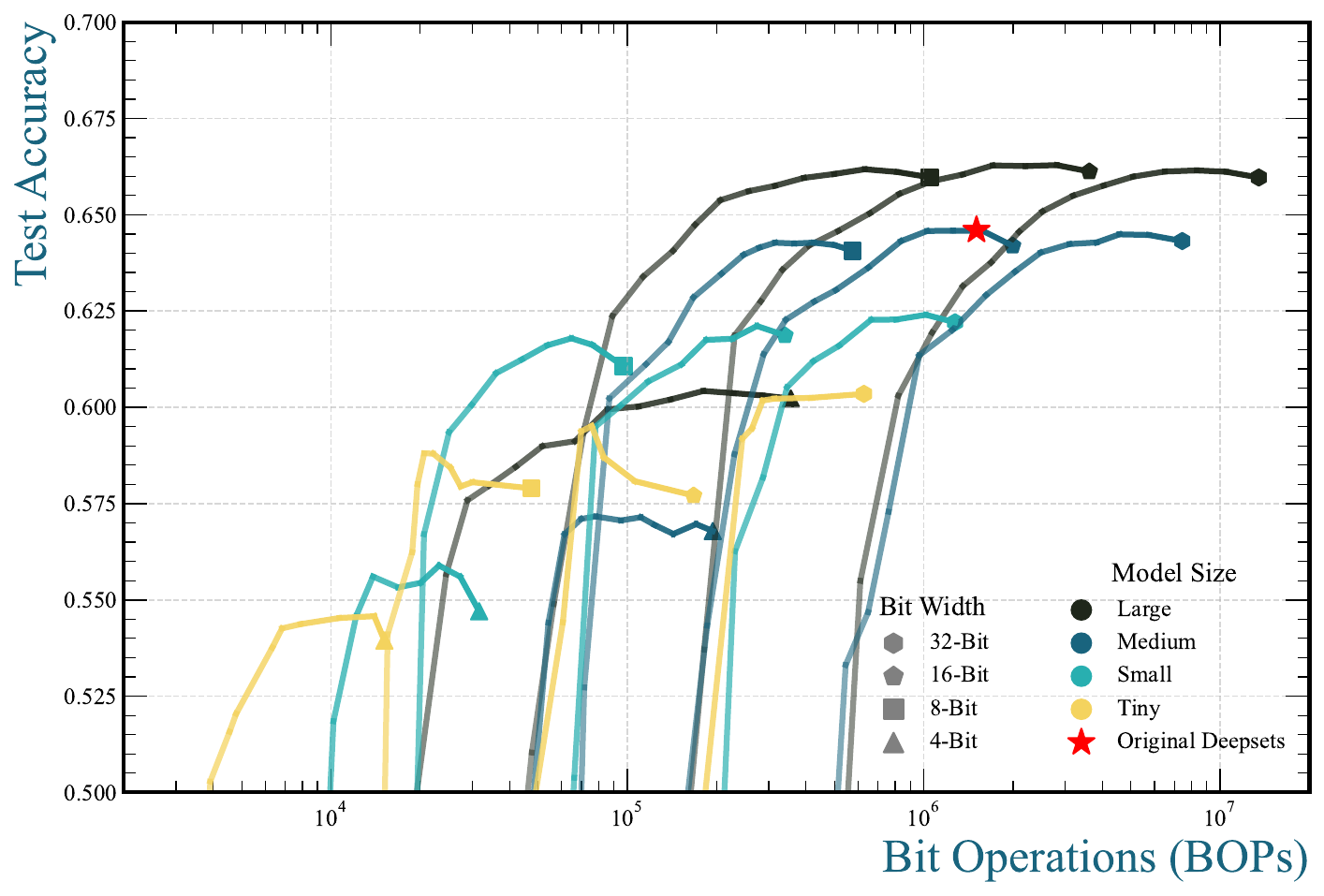}
 \caption{\label{fig:Deepsets pareto-front}Particle jet dataset Pareto-optimal front for the global (left) and local (right) searches. 
 The global search contains 2500 trials with the selected models in orange. 
 For the local search, each chosen model is quantized to various bit precisions, and then pruned by 20\% for 20 iterations indicated by increasing opacity.
 A bit width of 8 is chosen for its balance of BOPs and accuracy.}
\end{figure}

Our medium model achieved a 1.06\% increase in accuracy while reducing the number of BOPs by 7.2$\times$ and parameters by 23\% compared to the original deep sets model.
For applications where inference speed is critical, our tiny model has a compelling trade-off, with a 2.8\% decrease in accuracy from the original model but a substantial 30.25$\times$ reduction in BOPs and a 6$\times$ reduction in parameters.
Full model architectures can be accessed in Appendix Table~\ref{tab:particle_jet_models}.
Additionally, all models can be pruned to over 80\% sparsity (8 iterations) with less than a 10\% drop in performance, depicted in Fig.~\ref{fig:Deepsets pareto-front} (right).

When synthesized for FPGA deployment, our tiny model achieves a latency of only 70~ns with a 10~ns initialization interval, using less than 3\% of the available DSP slices, LUTs, and flip-flops on the target FPGA (Table~\ref{tab:Deepsets model_comparison2}).

The global search stage evaluated 1000 model architectures, utilizing 5 NVIDIA 3090 GPUs in parallel for a total of 61 hours. The subsequent local search stage trained 4 models with 4 quantization precisions each for 20 iterative magnitude pruning steps using 4 NVIDIA 3090 GPUs running in parallel for an additional 26 hours.


The performance of the deep sets models during global and local search can be seen in Fig.~\ref{fig:Deepsets pareto-front}.
As expected, accuracy drops down with increased pruning but the larger models can handle more sparisty before the performance drastically decreases.
Furthermore, the large model achieves a 2.5\% higher accuracy and a 0.87\% relative improvement in LUT utilization, although the inference time is slightly longer.



\section{Conclusion}

In this work, we have developed a pipeline for neural architecture codesign that streamlines the process of designing and optimizing deep learning models for physics applications.
Our method employs a two-stage search strategy, consisting of a global search to explore a wide range of architectures and a local search to fine-tune the most promising candidates.
We have demonstrated the effectiveness of our approach through two case studies: Bragg peak analysis for materials science and jet classification for particle physics.

For the BraggNN case study, our optimized models achieved comparable performance to the original model while significantly reducing computational complexity. 
Our large model improved the mean distance metric by 0.5\% while reducing BOPs by 5.9$\times$, and our small model increased mean distance by only 3\% while achieving a substantial 39.2$\times$ reduction in BOPs and 2.90$\times$ fewer parameters. When synthesized for FPGA deployment, our small model attained a latency of only 4.92 $\mu$s with less than 10\% utilization of DSPs, LUTs, and FFs.

In the particle jet classification task using the deep sets architecture, our optimized models again demonstrated significant improvements in efficiency with minimal impact on accuracy. 
The medium model increased accuracy by 1.06\% with a 7.2$\times$ reduction in BOPs and 23\% fewer parameters compared to the original.
For latency-critical applications, our tiny model trades a 2.8\% decrease in accuracy for a 30.25$\times$ reduction in BOPs and 6$\times$ fewer parameters.
Synthesized for target FPGA, it achieves an inference latency of only 70 ns while using less than 3\% of the FPGA resources.

The results from these case studies highlight the power of NAC to discover highly efficient ML architectures tailored for physics applications and resource-constrained edge devices.
This lowers the barrier to entry for domain experts looking to incorporate deep learning into their research.
By automating the neural architecture design and optimization process, we allow users to take advantage of state-of-the-art techniques without requiring extensive machine learning expertise.

While we exhibit a variety of architectures that either improve performance, latency, or resource utilization, the models found are limited in that none exceed in all criteria.
This limitation is primarily due to our BOPs metric that does not translate directly to latency, as it is only a general predictor for resources.
We plan for future work to cover more hardware-aware metrics, which can be done with proposed surrogate models~\cite{sherlock,rahimifar2024rule4mlopensourcetoolresource} to predict inference time.
OpenHLS~\cite{OpenHLS} is an example of another high level synthesis framework that synthesized the BraggNN model, achieving a latency of 4.8 $\mu$s.

Future work includes introducing more integrated parallelization techniques for each stage that will help dramatically decrease search time.
Additionally, our framework can be improved expanding the search space to incorporate more diverse layer types allowing for the discovery of more creative network architectures.
Investigating alternative search strategies tailored to specific physics tasks or FPGA synthesis should lead to further improvements.

\section*{Acknowledgments}
NT and JD are supported by the U.S. Department of Energy (DOE), Office of Science, Office of Advanced Scientific Computing Research under the ``Real‐time Data Reduction Codesign at the Extreme Edge for Science'' Project (DE-FOA-0002501).
JD is also supported by the DOE, Office of Science, Office of High Energy Physics Early Career Research program under Grant No. DE-SC0021187, and the U.S. National Science Foundation (NSF) Harnessing the Data Revolution (HDR) Institute for Accelerating AI Algorithms for Data Driven Discovery (A3D3) under Cooperative Agreement No. OAC-2117997.
NT is also supported by the DOE Early Career Research program under Award No. DE-FOA-0002019.

\appendix
\section{Bit Operations Calculation}
\label{sec: Bit Operations Calculation}

To evaluate the computational efficiency of our models, we calculate the bit operations (BOPs) for each layer.
The BOPs for linear and convolutional layers are calculated based on Ref.~\citep{BOP_Javi}, using the following equations.
For a linear layer with $m$ output features, $n$ input features, weight bit precision $b_w$, activation bit precision $b_a$, and sparsity $1-p$,
\begin{equation}
\mathrm{BOPs}(\mathrm{linear}) = m n (p b_a b_w + b_a + b_w + \log_2(n)).
\end{equation}
For a 2D convolutional layer with $m$ output features, $n$ input features, weight bit precision $b_w$, activation bit precision $b_a$, sparsity $1-p$, and kernel size $k$,
\begin{equation}
\mathrm{BOPs}(\mathrm{conv2d}) = m n k^2 (p b_a b_w + b_a + b_w + \log_2(nk^2)).
\end{equation}

To calculate the BOPs for the attention mechanism used in the BraggNN model, we first derive the BOPs for the softmax operation and matrix multiplication.
For softmax with input tensor of shape $(b, h, w, h, w)$ and bit precision $b_w$, the BOPs are
\begin{eqnarray}
\mathrm{BOPs}(\mathrm{softmax}) &=& (1.5)b(hw)^2 (b_w - 1)  + b hw (hw - 1) + b(hw)^2.
\end{eqnarray}
This accounts for the exponential function, summation, and division operations in softmax.

For matrix multiplication of matrices $\mathbf{A}$ and $\mathbf{B}$ with shapes $(b, m, n)$ and $(b, n, p)$, respectively,
\begin{equation}
\mathrm{BOPs}(\mathrm{matmul}) = b  m  n  (p b_w^2 + b_w (\log_2(n) + 1)).
\end{equation}
Combining these with the BOPs for 2D convolution, the total BOPs for our ConvAttention block with input shape $(b, c, h, w)$, hidden channels $d$, and kernel size $k$ is
\begin{eqnarray}
\mathrm{BOPs}(\mathrm{ConvAttn}) &=& \sum_i \mathrm{BOPs}({\mathrm{conv2d}_i}) + \mathrm{BOPs}(\mathrm{softmax}) \nonumber\\
&&+ \mathrm{BOPs}(\mathrm{matmul}_\mathrm{QK}) + \mathrm{BOPs}(\mathrm{matmul}_\mathrm{SV}),
\end{eqnarray}
where $i$ indexes over the $W_q$, $W_k$, $W_v$, and projection convolutions, QK indicates the query-key matrix multiplication with output shape $(b, hw, hw)$, and SV denotes the softmax-value matrix multiplication with output shape $(b, hw, d)$.

\section{Model Architectures}
\label{sec:model_arch}


\begin{table}[htpb]
\caption{\label{tab:braggnn_models}Architectures of optimized BraggNN models. 
$\mathrm{Conv}(a,b,c,d)$ denotes a 2D convolutional layer with $a$ input channels, $b$ output channels, kernel size $c$, and stride $d$.
BN denotes batch normalization layers.
LeakyReLU layers have a negative slope of $1/128$.}
\centering
\footnotesize
\begin{tabular}{@{}ll@{}}
\br
Model & Layers \\
\mr
Tiny & \begin{tabular}[c]{@{}l@{}}Conv(1, 8, 3, 1), Flatten, Linear($8\cdot9\cdot9$, 32), LeakyReLU,\\ Linear(32, 32), LeakyReLU, Linear(32, 32), BN, LeakyReLU,\\ Linear(32, 2), BN\end{tabular} \\
\mr
Small & \begin{tabular}[c]{@{}l@{}}Conv(1, 8, 3, 1), Conv(8,2,3,1), LeakyReLU, Conv(2,4,3,1), BN,\\ LeakyReLU, Flatten, Linear($4\cdot5\cdot5$, 64), BN, LeakyReLU,\\ Linear(64,32), BN, ReLU, Linear(32,16), BN, LeakyReLU,\\ Linear(16, 2), BN\end{tabular} \\
\mr
Medium & \begin{tabular}[c]{@{}l@{}}Conv(1, 8, 3, 1), Conv(8, 16, 3, 1), BN, ReLU, Conv(16, 4, 3, 1), BN,\\ LeakyReLU, Flatten, Linear($4\cdot5\cdot5$, 64), LeakyReLU, Linear(64, 64),\\ LeakyReLU, Linear(64, 16), BN, LeakyReLU, Linear(16, 2), BN\end{tabular} \\
\mr
Large & \begin{tabular}[c]{@{}l@{}}Conv(1, 8, 3, 1), Conv(8, 64, 3, 1), BN, Conv(64, 32, 3, 1), BN,\\ Flatten, Linear($5\cdot5\cdot32$, 32), LeakyReLU, Linear(32, 64), ReLU,\\ Linear(64, 64), BN, LeakyReLU, Linear(64, 2), BN\end{tabular} \\
\br
\end{tabular}
\end{table}

\begin{table}
\caption{\label{tab:particle_jet_models}Architectures of optimized particle jet models.
BN denotes batch normalization layers.
LeakyReLU layers have a negative slope of $1/128$.}
\centering
\footnotesize
\begin{tabular}{@{}lll@{}}
\br
Model & Component & Layers \\
\mr
\multirow{2}{*}{Tiny} & $\phi$ & Linear(3, 8), BN, ReLU, Linear(8, 8) \\
 & \multicolumn{2}{l}{\dotfill} \\
 & $\rho$ & Linear(8, 32), ReLU, Linear(32, 5) \\
\mr
\multirow{2}{*}{Small} & $\phi$ & Linear(3, 16), LeakyReLU, Linear(16, 8), BN \\
 & \multicolumn{2}{l}{\dotfill} \\
 & $\rho$ & \begin{tabular}[c]{@{}l@{}}Linear(8, 8), BN, ReLU, Linear(8, 16), LeakyReLU,\\ Linear(16, 16), LeakyReLU, Linear(16, 5), BN\end{tabular} \\
\mr
\multirow{2}{*}{Medium} & $\phi$ & Linear(3, 32), ReLU, Linear(32, 8) \\
 & \multicolumn{2}{l}{\dotfill} \\
 & $\rho$ & \begin{tabular}[c]{@{}l@{}}Linear(8, 32), BN, LeakyReLU, Linear(32, 16),\\ LeakyReLU, Linear(16, 64), BN, LeakyReLU, Linear(64, 5)\end{tabular} \\
\mr
\multirow{2}{*}{Large} & $\phi$ & Linear(3, 64), BN, ReLU, Linear(64, 16), BN \\
 & \multicolumn{2}{l}{\dotfill} \\
 & $\rho$ & \begin{tabular}[c]{@{}l@{}}Linear(16, 128), BN, ReLU, Linear(128, 16), BN,\\ LeakyReLU, Linear(16, 64), BN, ReLU, Linear(64, 5), BN\end{tabular} \\
\mr
\multirow{2}{*}{Original DeepSets} & $\phi$ & Linear(3, 32), ReLU, Linear(32, 32), ReLU \\
 & \multicolumn{2}{l}{\dotfill} \\
 & $\rho$ & Linear(32, 16), ReLU, Linear(16, 5) \\
\br
\end{tabular}
\end{table}

The model architectures found for the Bragg peak and particle jet classification task can be found in Tables ~\ref{tab:braggnn_models} and ~\ref{tab:particle_jet_models}.


\section{Search Spaces}

The search spaces for the BraggNN and particle jet models can be found in Tables~\ref{tab:BraggNN parameter_table} and~\ref{tab:DeepSets parameter_table}.

\begin{table}[htpb]
    \centering
    \small
    \renewcommand{\arraystretch}{1.2} 
    \caption{Comprehensive parameter space for BraggNN}
    \begin{tabular}{|l|l|}
    \hline
    \textbf{Parameter} & \textbf{Space or description} \\
    \hline
    \multicolumn{2}{|c|}{\textbf{General parameter space}} \\
    \hline
    Block & \{Conv, Attention, None\} \\
    Channel dimension & \{1, 2, 4, 8, 16, 32, 64\} \\
    Kernel size & \{1, 3\} \\
    Normalization method & \{Batch, None\} \\
    Activation function & \{ReLU, LeakyReLU, None\} \\
    Linear layer dimension & \{4, 8, 16, 32, 64\} \\
    \hline
    \multicolumn{2}{|c|}{\textbf{Conv block parameters}} \\
    \hline
    Conv1 in channels & Previous dimension \\
    Conv1 out channels & Sample channel dimension \\
    Conv1 kernel & Sample kernel size \\
    Norm1 & Sample normalization method \\
    Act1 & Sample activation function \\
    Conv2 in channels & Sample channel dimension \\
    Conv2 out channels & Sample channel dimension \\
    Conv2 kernel & Sample kernel size \\
    Norm2 & Sample normalization Method \\
    Act2 & Sample activation function \\
    \hline
    \multicolumn{2}{|c|}{\textbf{Attention block parameters}} \\
    \hline
    QKV Dimension & Sample channel dimension \\
    Skip Activation & Sample activation function \\
    \hline
    \multicolumn{2}{|c|}{\textbf{MLP classifier parameters} (all 4 layers)} \\
    \hline
    FC1 in dimension & Previous dimension \\
    FC1 out dimension & Sample linear space \\
    Norm1 & Sample normalization method \\
    Act1 & Sample activation function \\
    FC2 In Dimension & Previous dimension \\
    FC2 Out Dimension & Sample linear space \\
    Norm2 & Sample normalization method \\
    Act2 & Sample activation function \\
    \hline
    \end{tabular}
    \label{tab:BraggNN parameter_table}
\end{table}

\begin{table}[htpb]
    \centering
    \small
    \renewcommand{\arraystretch}{1.2} 
    \caption{Comprehensive parameter space for Deep Sets}
    \begin{tabular}{|l|l|}
    \hline
    \textbf{Parameter} & \textbf{Space or description} \\
    \hline
    \multicolumn{2}{|c|}{\textbf{$\phi$}} \\
    \hline
    Width & \{4, 8, 16, 32, 64\} \\
    Normalization method & \{Batch, None\} \\
    Activation function & \{ReLU, LeakyReLU, None\} \\
    Bottleneck dimension & \{1, 2, 4, 8, 16, 32, 64\} \\
    Aggregator & \{mean, maximum\} \\
    \hline
    \multicolumn{2}{|c|}{\textbf{$\rho$}} \\
    \hline
    Width & \{4, 8, 16, 32, 64\} \\
    Normalization method & \{Batch, None\} \\
    Activation function & \{ReLU, LeakyReLU, None\} \\
    \hline
    \end{tabular}
    \label{tab:DeepSets parameter_table}
\end{table}

\clearpage

\bibliographystyle{cms_unsrt}
\bibliography{biblio}

\end{document}